PAPER

# Attention from Above: A Multimodal Model for Drone-Based Object Localization

**Hyun-Ki Jung** (✉)

University of Seoul, Seoul, South Korea

stillhk3@uos.ac.kr

**ABSTRACT**

Drone-based object detection technology has advanced rapidly, becoming increasingly sophisticated and efficient. Recently, research trends have expanded beyond the detection of predefined objects toward the identification of specified target objects. For example, desired targets can be specified through textual prompts, enabling accurate detection of objects of interest. To address this demand, this paper proposes an efficient multimodal-based object detection model aimed at improving small object detection performance. The proposed method is built upon the YOLO-World framework and replaces the C2f layers used in the YOLOv8 backbone with attention-based A2C2f layers. This modification enables more precise representation of local features, particularly for small objects or objects with well-defined boundaries. In addition, the incorporation of attention mechanisms and parallel processing structures significantly enhances the model's computational accuracy. Comparative experiments conducted on the VisDrone dataset demonstrate that the proposed model outperforms the original YOLO-World model. Specifically, precision increases from 43.0% to 45.1%, recall from 32.8% to 35.0%, the F1 score from 37.2% to 39.4%, mAP@0.5 from 32.5% to 35.2%, and mAP@0.5-0.95 from 18.5% to 19.9%, confirming a substantial improvement in detection accuracy. These results verify that the proposed approach provides an effective and highly accurate solution for object detection in drone-based image and video application environments.

**KEYWORDS**

object detection, Yolo-World, text-guided, attention mechanisms, drone images

## 1 INTRODUCTION

Drones have evolved beyond simple recreational devices and are now widely utilized as professional tools across various fields, including photography and videography, inspection services in the manufacturing industry, and cargo and logistics transportation. In particular, advances in artificial intelligence (AI) have enabled autonomous flight and real-time data analysis, leading to the widespread adoption of drones and unmanned aerial vehicles (UAVs) across diverse industrial sectors.

Artificial intelligence significantly enhances the reliability and accuracy of drone technologies, thereby expanding their potential applications across a broad range of domains and industries [1]–[3].

Artificial intelligence-integrated drone-based object detection technologies have been applied in numerous fields and have driven substantial technological innovation. These technologies serve as a core foundation for industrial development and continue to be actively studied from diverse perspectives using multifaceted approaches. For example, Huang et al. designed a Unified Foreground Packing (UFP) module that clusters and merges sub-regions obtained from a dense detector [4]. Du et al. proposed a Global Context-Enhanced Adaptive Sparse Convolutional Network (CEASC) that leverages global contextual information to more efficiently cover foreground regions [5]. Li et al. introduced PointSR, a robust point-supervised object detection framework based on self-regularized sampling, which integrates temporal and informative constraints throughout the pseudo-box generation process [6]. Bi et al. proposed a Spatial Pyramid Pooling with Dilated Convolutions (SPPDSPC) module to strengthen feature extraction for densely packed targets [7].

Wu et al. incorporated a Multimodal Scale-Attentive Convolution (M-SC) module that dynamically generates modality-specific feature aggregation weights, enabling global cross-modality information fusion while reducing computational complexity [8]. Zhang et al. proposed a Hierarchical Consistency-based Active Learning (HiCAL) framework by constructing adversarially paired inputs under hierarchical perturbations [9]. Tschanz et al. evaluated the feasibility of combining drone-based image acquisition with deep learning to detect soil mounds representing bee nests and to distinguish them from other soil surface deposits, such as earthworm casts [10]. Zeng et al. proposed SPVD-DETR, a real-time end-to-end detector based on a Transformer architecture, contributing to the advancement of automated and intelligent detection of sweetpotato virus disease (SPVD) for large-scale, high-throughput phenotyping in precision agriculture [11]. Krzystek et al. analyzed the effects of tree species composition, stand density, and foliage condition on the robustness of deep-learning-based single-tree segmentation using UAV multispectral imagery and LiDAR-derived canopy height models (CHMs) [12]. Liu et al. designed the Group No-Local Attention Cross-Stage Partial Bottleneck with Two Convolutions (GNA-C2f) module and the Dual Branch Spatial Pyramid Pooling (DB-SPPF) module to enhance and fuse multi-scale features for traffic object detection from a UAV perspective [13]. Reshan et al. achieved significant advancements in preventive bushfire management by utilizing UAVs, improving cost efficiency in real-world environments through high-quality monitoring and precise detection of fire indicators [14].

In the future, drone-based image and video object detection technologies should evolve beyond conventional computer vision (CV)-centric approaches that merely focus on object detection toward multimodal models that integrate natural language processing (NLP) and large language models (LLMs). Through such integration, user-specified target objects are expected to be detected more accurately and efficiently, even in complex environments.

- The main contributions of this paper are summarized as follows:
  1. The primary objective of this paper is to propose a text-guided object detection model capable of accurately detecting small objects. To achieve this goal, extensive experiments were conducted using the VisDrone dataset, which consists of drone-captured images collected under diverse environmental conditions. Based on these experiments, a multimodal object detection model optimized for small object detection was developed.

2. This study proposes a novel network architecture in which the C2f layers of the original YOLO-World backbone are replaced with attention-based A2C2f layers. As a result, performance improvements were observed across multiple evaluation metrics, including precision, recall, F1 score, mAP@0.5, and mAP@0.5-0.95. In particular, the mAP@0.5 increased from 32.5 to 35.2, confirming a substantial improvement in detection accuracy. These results demonstrate that the proposed model achieves superior performance in terms of detection accuracy.

# 2 RELATED WORKS

## 2.1 Object detection models

Object detection models are generally classified into two main categories: one-stage detectors and two-stage detectors. Two-stage detectors perform object detection through a sequential two-step process. In the first stage, candidate regions that are likely to contain objects are generated. In the second stage, these regions are analyzed in greater detail to determine precise object locations and the corresponding target classes. To generate candidate regions, techniques such as selective search and the sliding window method are commonly employed. Selective search identifies potential object regions by grouping similar pixels, whereas the sliding window method sequentially scans the entire image or video using fixed-size rectangular windows to extract object candidates. Due to this multi-step procedure, two-stage detectors generally involve a more complex detection pipeline than one-stage detectors. Representative two-stage detectors include R-CNN, Fast R-CNN, Faster R-CNN, and Mask R-CNN [15]–[18], which have been widely adopted in the field of computer vision.

In contrast, one-stage detectors perform region proposal and classification simultaneously using convolutional neural networks. Because all tasks are processed in parallel within the convolutional layers responsible for feature extraction, one-stage detectors achieve significantly faster detection speeds than two-stage detectors. Moreover, owing to their relatively simple training and inference processes, they have been increasingly applied across a wide range of application domains in recent years. Representative models in this category include the You Only Look Once (YOLO) family [19]–[31], the Single Shot MultiBox Detector (SSD), Focal Loss–based detectors, and RefineDet [32]–[34].

In this study, the backbone network of the YOLO model, which belongs to the one-stage detector category, is modified and employed for feature extraction. Specifically, performance improvements are achieved by enhancing the YOLOv8 backbone used in the YOLO-World model.

## 2.2 Vision-language model

Autoregressive multimodal vision–language models (AVLMs) have emerged as promising alternatives, demonstrating outstanding visual reasoning performance across various domains [35]. Multimodal approaches associated with such vision–language models have been actively investigated in a wide range of fields. Pu et al. proposed a novel approach to improve image–text matching performance through a graph-based aggregation–disentanglement framework (GADNet) [36]. Li et al. constructed a multimodal image–text retrieval system based on the fusion of pre-trained models by increasing the diversity of the original data using the MixGen algorithm [37]. Sun et al. proposed a novel method, termed Strong and Weak Prompt Engineering (SWPE), for remote sensing image–text retrieval [38]. Zhang et al. developed a local-enhanced representation framework by designing a relation-based cross-modal local-enhanced fusion module to suppress noise in local features [39]. Xiao et al. investigated a novel approach, termed "Fine-grained Language-informed Image Representations (FLAIR)," to capture more fine-grained visual features [40]. Binyamin et al. described a method called CountGen, which generates prompt- and seed-dependent layouts by leveraging the diffusion model's own prior rather than relying on external information to determine object layouts [41]. Huang et al. introduced a large language model (LLM)-based perception-guided jailbreak (PGJ) method, inspired by the observation that semantically different texts can lead to similar human perceptions [42].

The YOLO-World model used as the baseline in this study is a multimodal framework that extends the conventional YOLO architecture with open-vocabulary object detection capabilities through vision–language modeling and large-scale pre-training on CLIP datasets. The original YOLO-World model combines the efficient detection performance of the YOLOv8 backbone with the strong textual understanding and cross-modal reasoning capabilities of CLIP. In this study, an attention mechanism is further incorporated into the CLIP-based YOLO-World model to improve detection accuracy.

## 3 METHODOLOGY

### 3.1 The proposed model

The overall architecture of the YOLO-World model integrated with the backbone network proposed in this paper is illustrated in detail in Figure 1. The input image shown in Figure 1 is an actual sample from the VisDrone dataset used in the experiments [43].

To describe the overall architecture, the proposed YOLO-World model first receives a user-provided prompt sentence that includes terms such as "car" and "truck." The text encoder then converts the input text into embeddings using the frozen weights of the CLIP model, while the image encoder encodes the input image into multi-scale visual features based on the modified YOLOv8 backbone, which is highlighted in red in Figure 1. Subsequently, the re-parameterizable vision-language PAN (RepVL-PAN) performs multi-level cross-modal fusion on the image and text features. Finally, the proposed YOLO-World model predicts regressed bounding boxes and object embeddings corresponding to the nouns or descriptions contained in the input sentence.

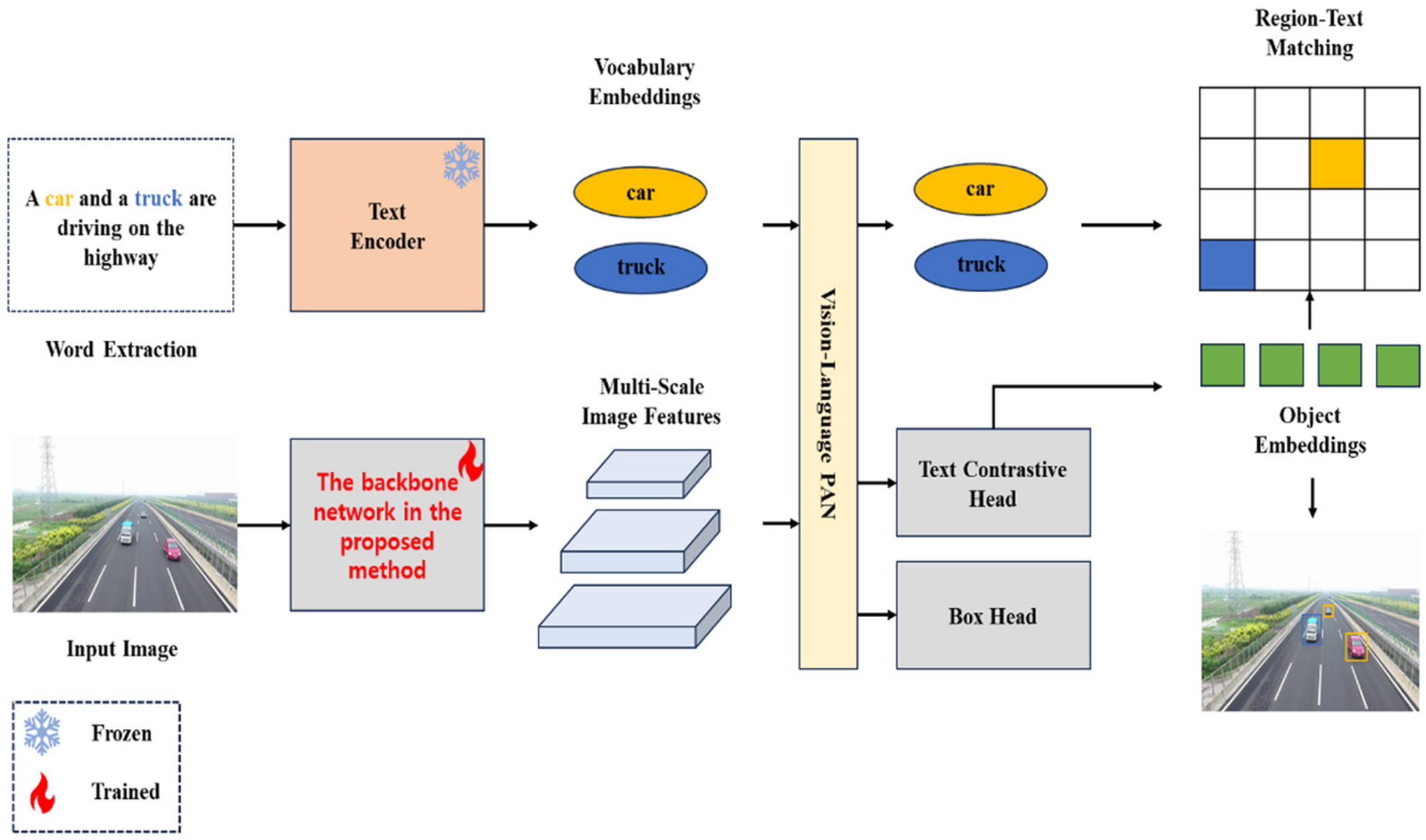

**Fig. 1.** Overall architecture of the proposed model

## 3.2 Optimized YOLO backbone architecture

The original YOLO-World model adopts the C2f layers used in the YOLOv8 backbone network as its core building blocks. In contrast, this study constructs an enhanced backbone network by replacing the conventional C2f layers with A2C2f layers that integrate an attention mechanism as the core component. The A2C2f block incorporates an area attention module (A2) into the C2f structure, thereby improving feature discrimination while preserving computational efficiency.

The A2 module serves as a fundamental component of the A2C2f layer by recalibrating feature responses through adaptive attention aggregation to emphasize informative regions while effectively suppressing irrelevant background information. In other words, the A2 module is a simple yet efficient area-attention mechanism that maintains a large receptive field while significantly reducing the computational complexity of attention operations, thereby enhancing processing speed [44].

As illustrated in Figure 2a, the A2C2f layer parallelizes ABlock layers to integrate multi-head MLP blocks with localized area attention mechanisms. This design enhances spatial feature learning capability while maintaining lightweight computational complexity.

Furthermore, as shown in Figure 2b, the C2PSA (Cross Stage Partial with Spatial Attention) module extends the C2f block of the YOLOv8 family by incorporating a spatial attention (SA) mechanism. While preserving the original CSP/C2f channel structure, this module is designed to explicitly learn the importance of spatial locations, thereby significantly enhancing spatial attention within feature maps [45]. Through this spatial attention mechanism, the model can more effectively focus on salient regions in the image. By spatially pooling feature information, the C2PSA layer improves detection accuracy for objects of varying scales and locations.

Finally, Figure 2c illustrates the overall architecture of the proposed YOLO-based backbone network. The modified layers described in Figure 2a and 2b are highlighted in red.

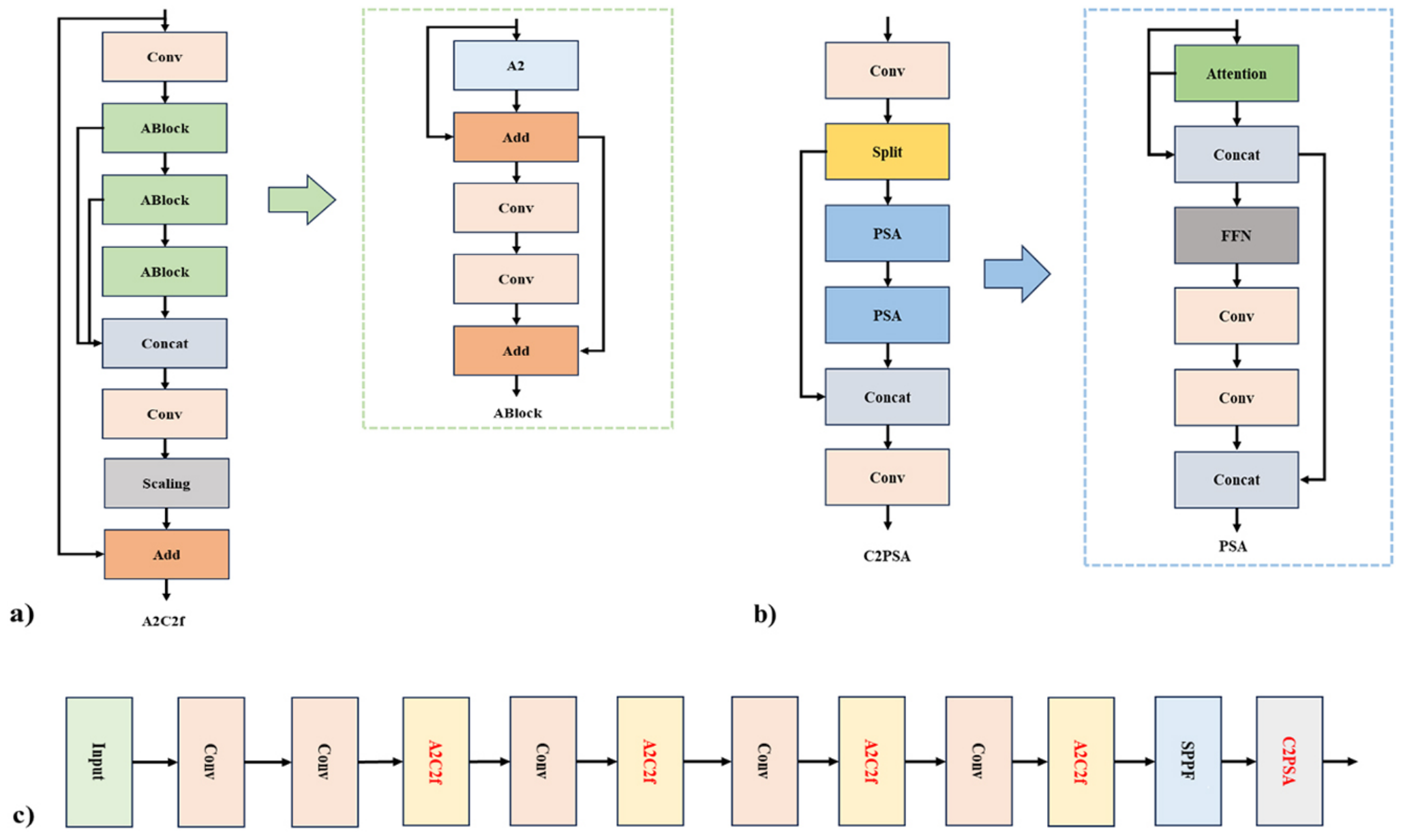

**Fig. 2.** Architecture of the backbone network in the proposed YOLO-World model: (a) A2C2f layer flowchart, (b) C2PSA layer flowchart, (c) Modified YOLO backbone network

## 3.3 Evaluation metrics

In this paper, precision (P), recall (R), F1 score, average precision (AP), and mean average precision (mAP) are employed as the primary metrics to evaluate and compare all experimental results. The detailed definitions and formulations of these metrics are provided in Equations (1)–(5), and the key terms used in these equations are defined as follows. A true positive (TP) refers to a case in which the model correctly identifies a positive instance. A false positive (FP) occurs when the model incorrectly classifies a negative instance as positive. A false negative (FN) denotes a case in which the model fails to detect a positive instance and instead classifies it as negative. Finally, a true negative (TN) represents a case in which the model correctly identifies a negative instance as negative.

Based on these definitions, precision (P), recall (R), and the F1 score are defined. Precision is the proportion of true positives among all instances classified as positive by the model. For example, in the context of vehicle identification in images, precision indicates the proportion of predicted vehicles that are actually vehicles. Recall is defined as the proportion of true positives among all actual positive instances, representing the fraction of real vehicles that are successfully detected by the model. The F1 score, derived from precision and recall, is defined as their harmonic mean. Since a trade-off generally exists between precision and recall, the F1 score serves as a meaningful metric for evaluating model performance, particularly in scenarios involving data imbalance.

$$P = \frac{TP}{TP + FP} \tag{1}$$

$$R = \frac{TP}{TP + FN} \tag{2}$$

$$F1score = \frac{2 \times Precision \times Recall}{Precision + Recall} \tag{3}$$

To analyze the trade-off between precision and recall, the precision–recall (PR) curve is employed, as it enables an accurate evaluation of metrics that exhibit such conflicting characteristics. Based on the PR curve, average precision (AP) quantifies the performance of an object detection algorithm by computing the area under the curve, thereby summarizing detection accuracy as a single scalar value. A higher AP value indicates superior model accuracy. In this paper, mean average precision (mAP), obtained by averaging the AP values across all classes, is adopted as the final metric for evaluating overall model performance.

$$AP = \int_0^1 P(r)dr \tag{4}$$

$$mAP = \frac{1}{N}\sum_{i=1}^{N} AP_i \tag{5}$$

# 4 RESULTS AND DISCUSSION

## 4.1 Experimental setup

As shown in Table 1, the baseline experimental environment in this study was configured using Google Colab. Python (version 3.12.12) was used as the programming language, and PyTorch (version 2.9.0) was employed as the deep learning framework. The experiments were conducted on an NVIDIA Tesla T4 GPU supporting CUDA 12.6, and the system was equipped with 15 GB of RAM. The number of training epochs was set to 100, and the initial input resolution of the model was set to 640 × 640 pixels. In addition, all experiments were performed based on the default configuration of YOLOv8n, with the depth multiple set to 0.33 and the width multiple set to 0.25.

**Table 1.** Hardware and software configuration

| Name | Parameters |
|---|---|
| Development environment | Google Colab |
| GPU | Nvidia, Tesla T4 |
| Installed RAM | 15GB |
| CUDA Version | 12.6 |
| Programming language | Python 3.12.12 |
| Deep learning framework | PyTorch 2.9.0 |

In this paper, the VisDrone dataset is used for performance comparison experiments with the existing YOLO-World model. VisDrone is one of the most widely used benchmark datasets in the field of drone-based object detection and is commonly employed to evaluate and compare the performance of object detection models using images and videos captured by drones. The dataset consists of a total of 8,629 images, including 6,471 images for training, 548 for validation, and 1,610 for testing. The objects to be detected are categorized into ten classes: pedestrian, people, bicycle, car, van, truck, tricycle, awning tricycle, bus, and motor, all of which are captured in diverse real-world environments. Example images from the dataset used in the experiments are presented in Figure 3.

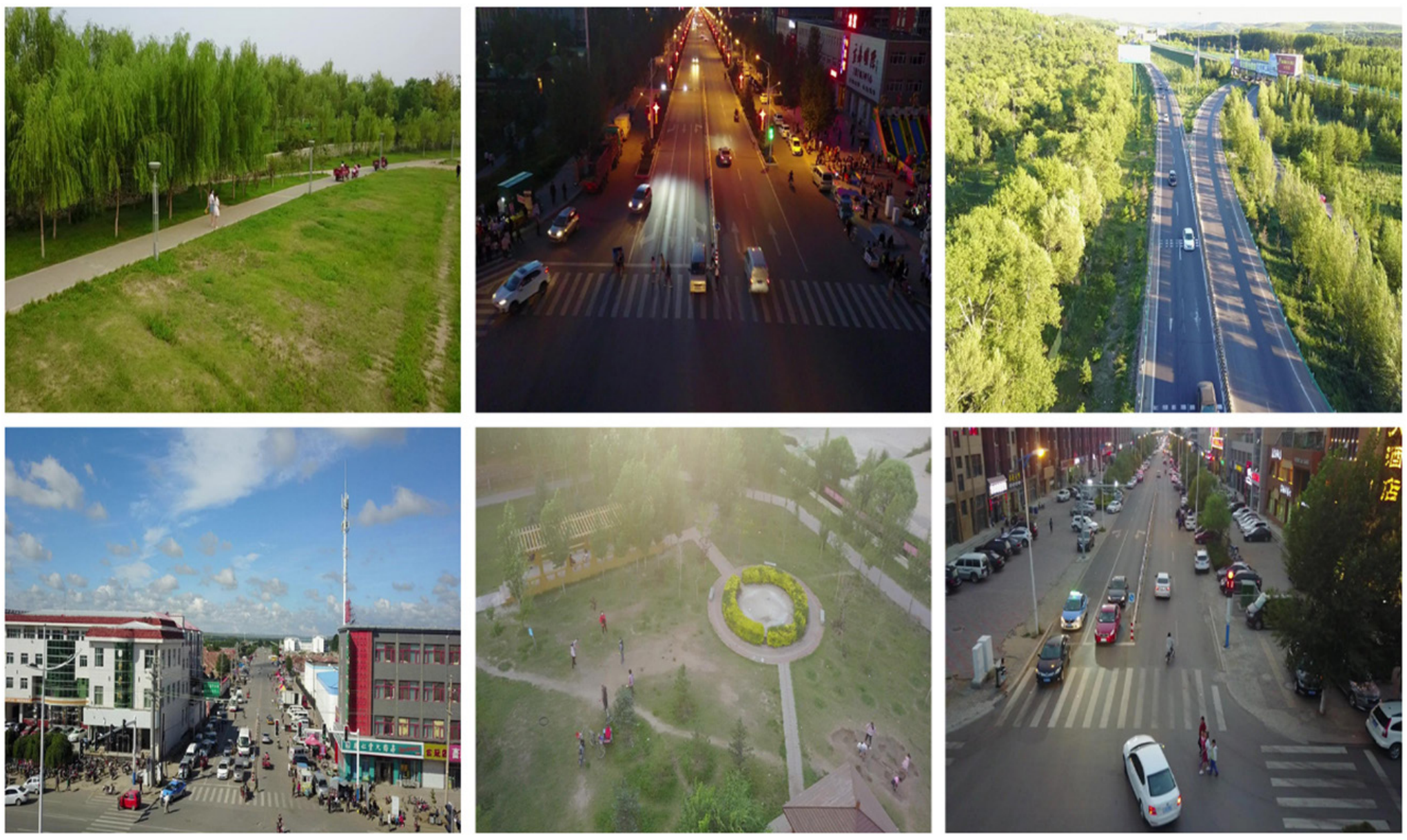

**Fig. 3.** Representative images from the VisDrone dataset

## 4.2 Experimental results

Figure 4 illustrates the types, quantities, and distributions of all object classes in the VisDrone dataset used in the experiments with the proposed model. Figure 4a presents the name of each object class along with the corresponding number of instances, demonstrating that the VisDrone dataset contains a sufficient number of samples for conducting the experiments. Figure 4b shows the distribution of object labels, where the x-axis represents the ratio of the label center to the image width and the y-axis represents the ratio of the label center to the image height. This distribution indicates that the labels are generally well distributed, with a relatively high concentration near the center of the images. Figure 4c depicts the size distribution of each object class, where the x-axis denotes the ratio of the label width to the image width and the y-axis denotes the ratio of the label height to the image height.

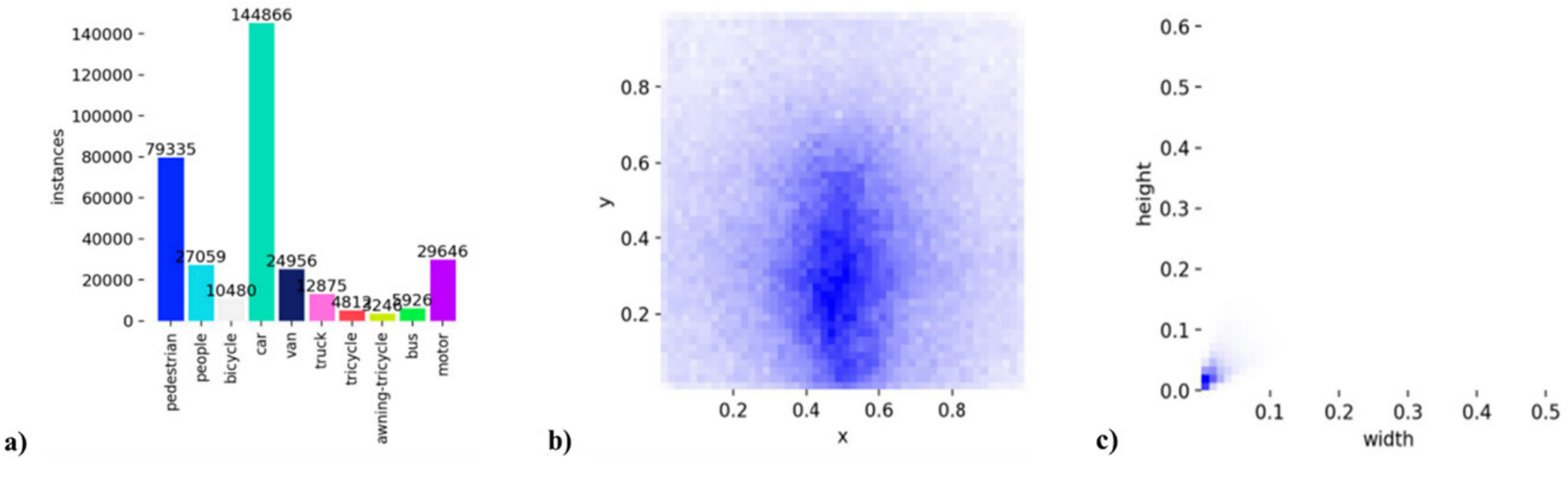

**Fig. 4.** Instance counts and label distributions across classes: (a) instance counts, (b) label positions, (c) label sizes

The experiments in this paper were conducted independently using both the original YOLO-World model and a modified version incorporating an improved backbone network. For comparative analysis, the proposed model replaces the C2f layers in the original YOLO-World backbone with attention-based A2C2f layers. To evaluate object detection performance, standard metrics including precision, recall, F1 score, mAP@0.5, and mAP@0.5-0.95 were employed.

The experimental results demonstrate that, compared with the original YOLO-World model, precision increased from 43.0% to 45.1%, representing an improvement of 2.1%, while recall rose from 32.8% to 35.0%, yielding a gain of 2.2%. The F1 score improved from 37.2% to 39.4%, corresponding to an increase of 2.2%. In addition, mAP@0.5 increased from 32.5% to 35.2%, reflecting a 2.7% improvement, and mAP@0.5-0.95 increased from 18.5% to 19.9%, indicating a substantial enhancement in overall detection performance across multiple IoU thresholds.

Furthermore, performance comparison experiments were conducted against state-of-the-art (SOTA) object detection models, including YOLOv9, YOLOv10, YOLOv11, YOLOv12, and YOLOv26, with the results summarized in Table 2. The class-wise detection results obtained using the modified model are presented in Table 3. As shown in Table 3, relatively low mAP@0.5 values were observed for the bicycle, tricycle, and awning-tricycle classes. This performance degradation can be attributed to the small object sizes, which make these classes difficult to distinguish, as well as their limited representation in the dataset. In other words, the lower mAP@0.5 and mAP@0.5-0.95 scores for these categories indicate a higher level of detection difficulty compared with other object classes.

**Table 2.** Performance comparison of various object detection models

| Method | Vision Backbone | Text Model | Precision (%) | Recall (%) | F1 Score (%) | mAP@0.5 (%) | mAP@0.5–0.95 (%) |
|---|---|---|---|---|---|---|---|
| YOLOv9 | YOLOv9 | – | 44.6 | 32.6 | 37.7 | 33.0 | 18.8 |
| YOLOv10 | YOLOv10 | – | 43.7 | 31.9 | 36.9 | 31.8 | 18.2 |
| YOLOv11 | YOLOv11 | – | 41.7 | 32.2 | 36.3 | 31.7 | 18.2 |
| YOLOv12 | YOLOv12 | – | 43.5 | 31.8 | 36.7 | 32.1 | 18.4 |
| YOLOv26 | YOLOv26 | – | 41.2 | 32.6 | 36.4 | 31.8 | 18.2 |
| YOLO-World | YOLOv8 | CLIP | 43.0 | 32.8 | 37.2 | 32.5 | 18.5 |
| Proposed Model | Proposed Backbone | CLIP | 45.1 | 35.0 | 39.4 | 35.2 | 19.9 |

**Table 3.** Detection results by object category in the proposed model

| Class | Instances | Precision (%) | Recall (%) | F1 Score (%) | mAP@0.5 (%) | mAP@0.5–0.95 (%) |
|---|---|---|---|---|---|---|
| All | 38759 | 45.1 | 35.0 | 39.4 | 35.2 | 19.9 |
| Pedestrian | 8844 | 46.2 | 39.1 | 42.4 | 38.9 | 16.3 |
| People | 5125 | 49.1 | 27.7 | 35.4 | 30.5 | 10.9 |
| Bicycle | 1287 | 25.5 | 10.3 | 14.7 | 10.4 | 3.9 |
| Car | 14064 | 65.7 | 76.6 | 70.7 | 77.4 | 52.8 |
| Van | 1975 | 48.3 | 41.3 | 44.5 | 42.0 | 28.6 |
| Truck | 750 | 44.1 | 31.2 | 36.5 | 30.1 | 18.8 |
| Tricycle | 1045 | 37.8 | 23.7 | 29.1 | 22.7 | 11.7 |
| Awning-Tricycle | 532 | 33.1 | 16.2 | 21.8 | 14.7 | 9.0 |
| Bus | 251 | 54.3 | 43.0 | 48.0 | 46.2 | 30.6 |
| Motor | 4886 | 47.2 | 41.2 | 44.0 | 39.4 | 16.1 |

The graph shown on the left side of Figure 5a presents the class-wise precision–recall curves of the proposed model. Among the classes used in the experiment, the bicycle class recorded the lowest value of 0.104, while the awning-tricycle class showed the second-lowest value at 0.147. In contrast, the car class achieved the highest value of 0.744. The average precision–recall value across all classes was confirmed to be 0.352.

Meanwhile, the graph on the right side of Figure 5b illustrates the mAP@0.5 and loss curves over 100 training epochs, indicating that the proposed model converged stably and effectively. As an additional experiment, identical sentences were separately provided as inputs to both the original YOLO-World model and the modified model. Each sentence included the classes motor, van, people, awning-tricycle, and truck, which were used as prompts in four respective cases. The comparative results of these experiments are presented in Table 4.

a)
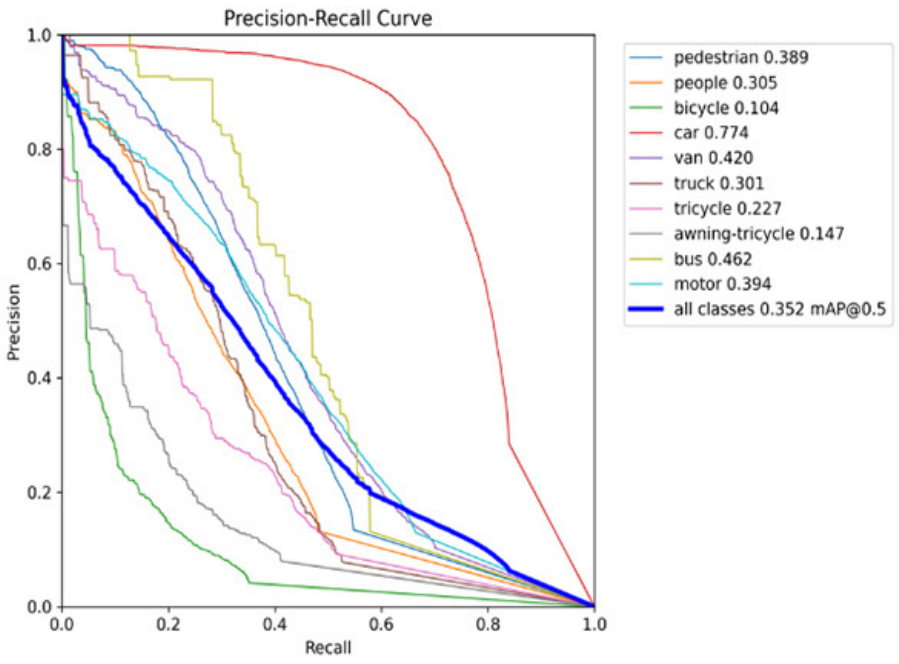

b)
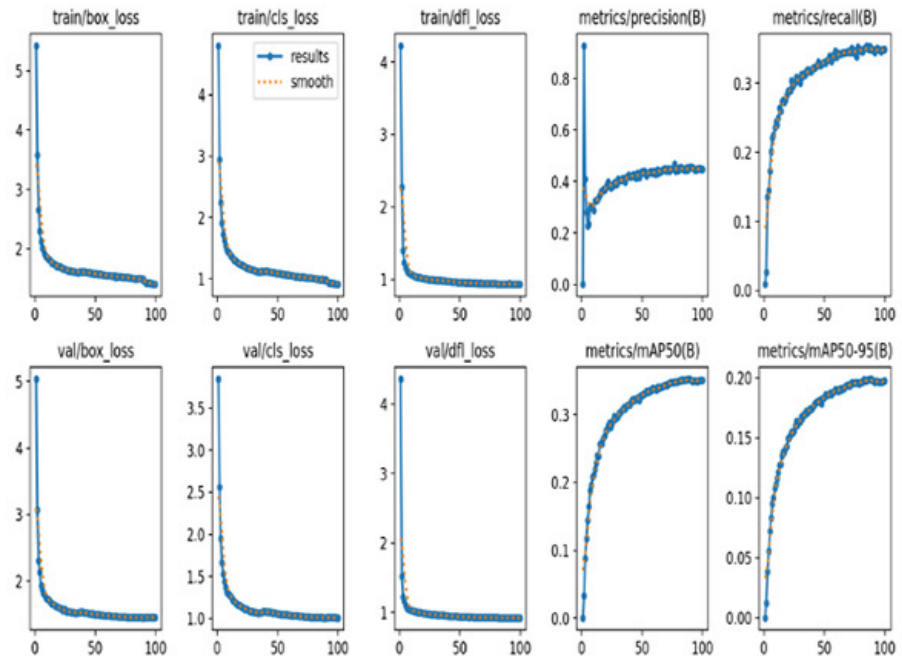

**Fig. 5.** Performance evaluation of the proposed model: (a) precision–recall curves, (b) evolution of key metrics during training

**Table 4.** Comparison of object detection results by command according to prompt text input

| Methods | Command 1: I Want to Know Where the Motor is | Command 2: Show Me the Location of the Van | Command 3: Tell Me Where the People are | Command 4: Locate the Awning-Tricycle and the Truck |
|---|---|---|---|---|
| YOLO-World | 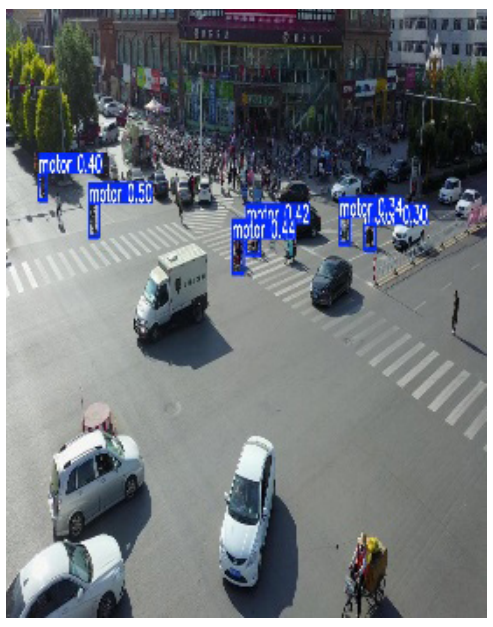 | 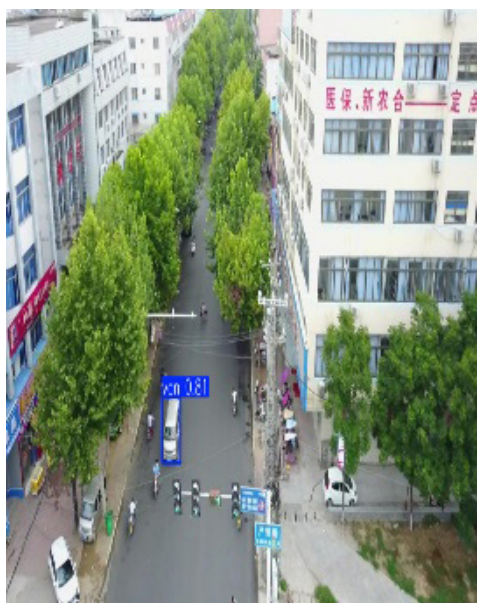 | 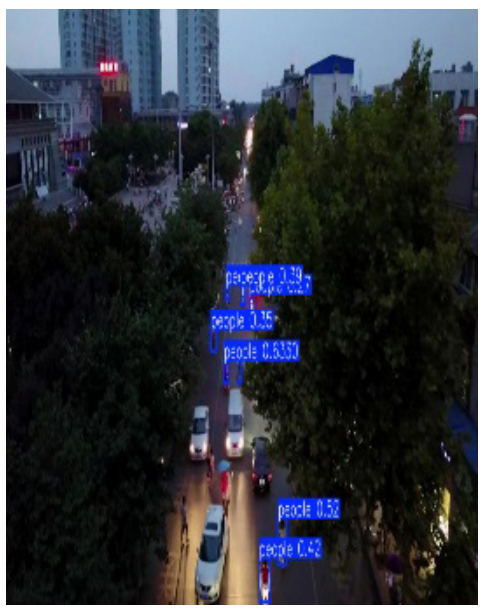 | 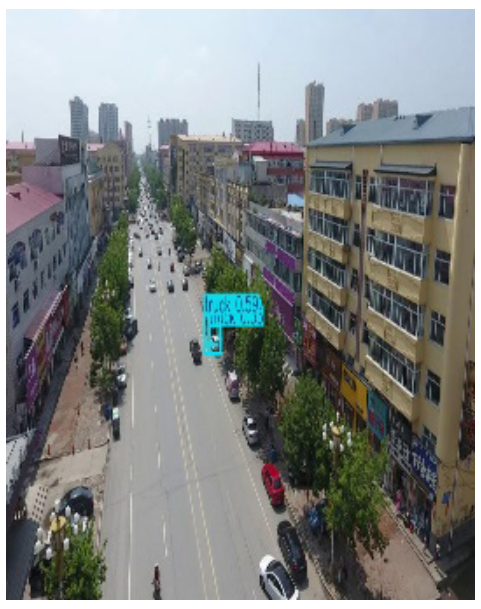 |
| Proposed Model | 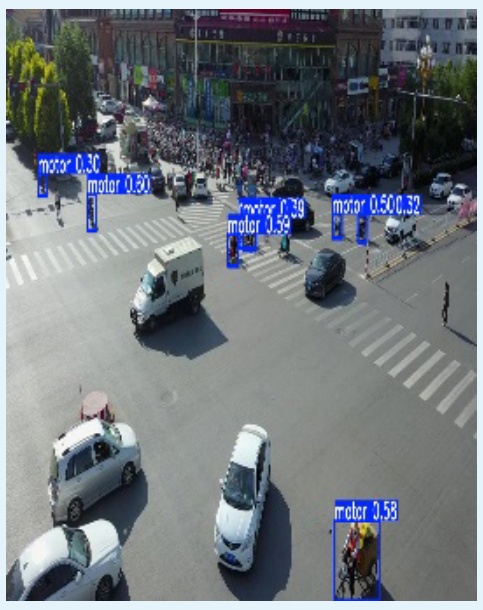 | 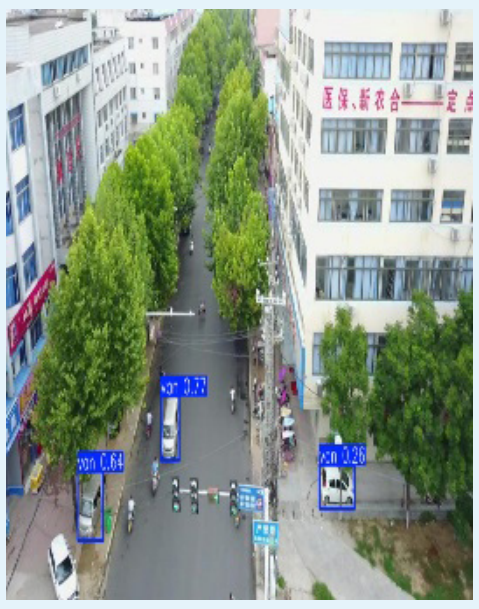 | 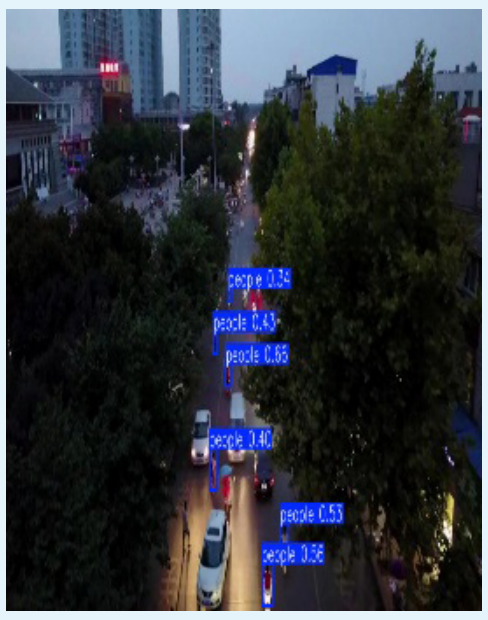 | 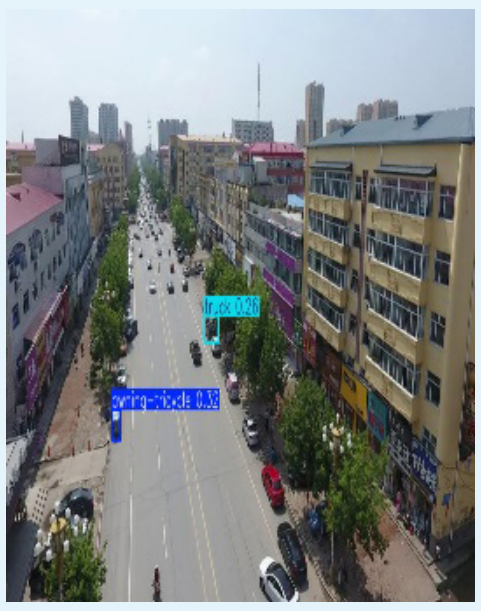 |

## 5 CONCLUSIONS

This paper proposes an extended text-guided object detection model for accurately detecting small objects in drone images and videos, building upon previous research [46]. The proposed approach focuses on improving detection accuracy by enhancing the YOLOv8 backbone network used for feature extraction in the original YOLO-World model. To validate the effectiveness of the proposed method, a series of experiments was conducted. The experimental dataset was constructed based on the widely used VisDrone dataset, which includes ten object classes: pedestrian, people, bicycle, car, van, truck, tricycle, awning-tricycle, bus, and motor.

The proposed backbone network replaces the conventional C2f layers with attention-based A2C2f layers, introducing a new backbone architecture designed to improve detection accuracy. Experimental results demonstrate that the proposed method outperforms the original YOLO-World model in terms of detection performance, achieving mAP@0.5 and mAP@0.5-0.95 scores of 30.7% and 19.9%, respectively. These results correspond to improvements of 0.3% and 0.4% over the original YOLO-World model. Overall, the results confirm that the proposed model achieves higher accuracy in small object detection. Consequently, the findings of this study are expected to provide effective technical solutions for a wide range of industrial and academic applications involving drone-based vision systems.

## 6 FUNDING INFORMATION

This study received no external funding.

## 7 CONFLICTS OF INTEREST

The author declares that there are no conflicts of interest to report regarding the present study.

## 8 DATA AVAILABILITY STATEMENT

All the data used in the experiments was based on the VisDrone: 2019 dataset.